\documentclass[letterpaper]{article} 
\usepackage{aaai23}  
\usepackage{times}  
\usepackage{helvet}  
\usepackage{courier}  
\usepackage[hyphens]{url}  
\usepackage{graphicx} 
\def\UrlFont{\rm}  
\usepackage{natbib}  
\usepackage{caption} 
\usepackage{algorithm}
\usepackage{algorithmic}
\usepackage{newfloat}
\usepackage{listings}
\DeclareCaptionStyle{ruled}{labelfont=normalfont,labelsep=colon,strut=off} 
\floatstyle{ruled}
\newfloat{listing}{tb}{lst}{}
\floatname{listing}{Listing}
\usepackage[algo2e,ruled,vlined]{algorithm2e}
\usepackage[table,xcdraw]{xcolor}
\usepackage{amsmath}
\usepackage{amssymb}
\usepackage{multirow}
\usepackage{adjustbox}
\usepackage{threeparttable}
\usepackage{algorithmic}

\usepackage{newfloat}
\usepackage{listings}
\usepackage{booktabs}
\floatstyle{ruled}
\newfloat{listing}{tb}{lst}{}
\floatname{listing}{Listing}
\DeclareCaptionStyle{ruled}{labelfont=normalfont,labelsep=colon,strut=off} 
\title{Revisiting the Spatial and Temporal Modeling for Few-Shot Action Recognition}

\author{
Jiazheng Xing,  Mengmeng Wang\footnote{Co-corresponding author.}, Yong Liu\footnote{Corresponding author.}, Boyu Mu
}

\affiliations{
    \textsuperscript{\rm} Zhejiang University, Hangzhou, China \\

    \{jiazhengxing,mengmengwang, muboyu\}@zju.edu.cn, yongliu@iipc.zju.edu.cn
}
\usepackage{bibentry}

\begin{document}
	
	\maketitle
	
	\begin{abstract}
        Spatial and temporal modeling is one of the most core aspects of few-shot action recognition. Most previous works mainly focus on long-term temporal relation modeling based on high-level spatial representations, without considering the crucial low-level spatial features and short-term temporal relations. Actually, the former feature could bring rich local semantic information, and the latter feature could represent motion characteristics of adjacent frames, respectively. In this paper, we propose SloshNet, a new framework that revisits the spatial and temporal modeling for few-shot action recognition in a finer manner. First, to exploit the low-level spatial features, we design a feature fusion architecture search module to automatically search for the best combination of the low-level and high-level spatial features. Next, inspired by the recent transformer, we introduce a long-term temporal modeling module to model the global temporal relations based on the extracted spatial appearance features. Meanwhile, we design another short-term temporal modeling module to encode the motion characteristics between adjacent frame representations. After that, the final predictions can be obtained by feeding the embedded rich spatial-temporal features to a common frame-level class prototype matcher. We extensively validate the proposed SloshNet on four few-shot action recognition datasets, including Something-Something V2, Kinetics, UCF101, and HMDB51. It achieves favorable results against state-of-the-art methods in all datasets.
 
	\end{abstract}
	
	\section{Introduction}
	\begin{figure} [ht]
		\centering
		\includegraphics[width=\linewidth,height=0.9\linewidth]{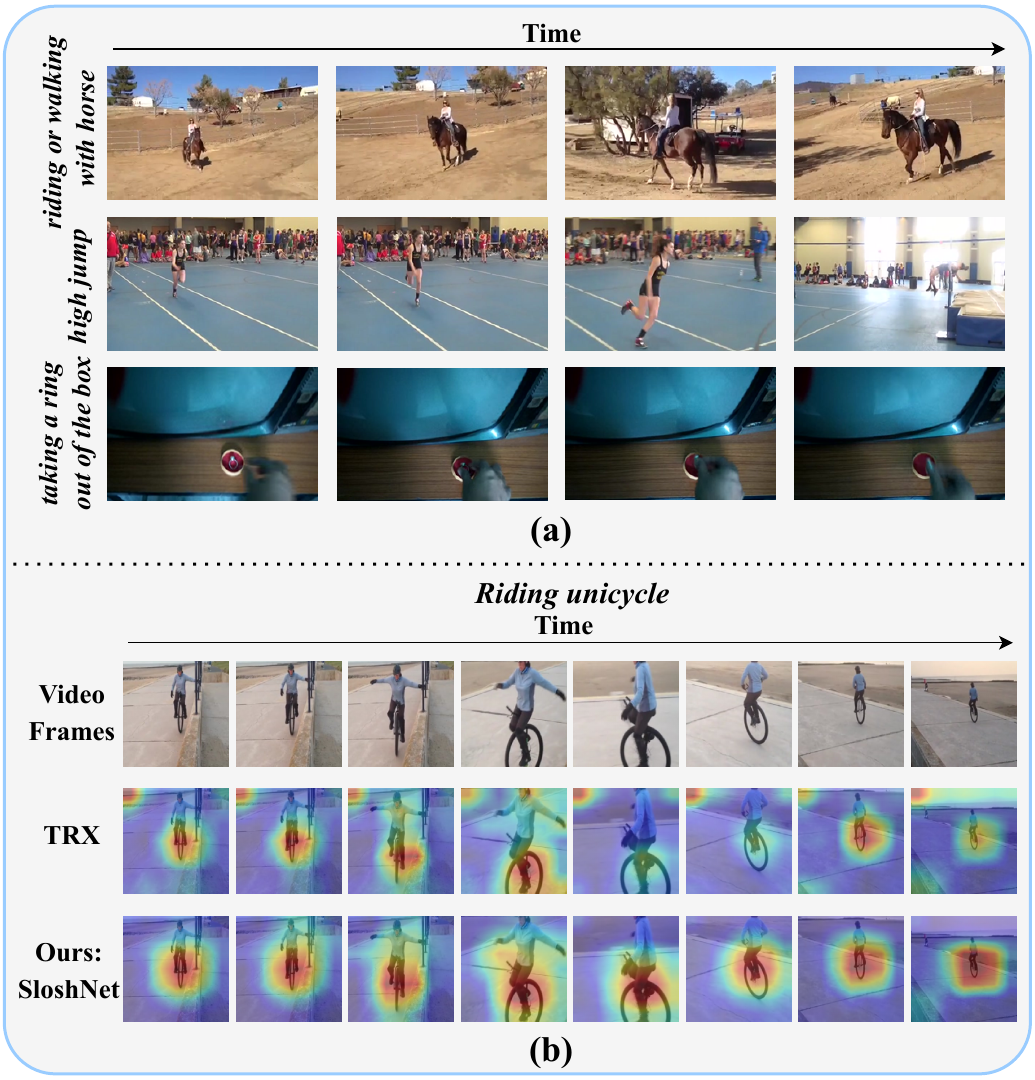}
		\caption{(a): Some examples in few-shot action recognition (b): Visualization of the attention map from the recent work TRX~\cite{perrett2021temporal} and our proposed SloshNet.}
		\label{fig:fig1}
        \vspace{-10pt}
	\end{figure}
	With the development of deep learning, a large amount of excellent work has emerged in the field of action recognition~\cite{li2022uniformer,liu2022video,wang2022learning,feichtenhofer2019slowfast, wang2018temporal}. Most studies use large amounts of labeled data to perform video understanding or classification tasks to learn video representations. Such approaches are unsatisfactory in industrial applications because of the massive time-consuming and labor-consuming data annotation. On the contrary, the core assumption of few-shot learning is using only a handful of labeled training samples from numerous similar tasks as a surrogate for large quantities of labeled training samples. Therefore, the attention on few-shot learning methods is increasing daily.
	The task of few-shot action recognition aims to classify an unlabeled query video into one of the action categories in the support set (usually five categories) with limited samples per action class. 
	
	Inspired by few-shot image recognition ~\cite{finn2017model, doersch2020crosstransformers, elsken2020meta,ma2020transductive}, existing few-shot video action recognition methods mainly focus on comparing the similarity of different videos in the feature space for recognition. However, videos have an extra temporal dimension compared to images, so that it is insufficient to represent the whole video as a single feature vector. Therefore, the spatial-temporal feature modeling becomes one of the core problems of few-shot action recognition. Specially, spatial feature aims to express spatial semantic information for every single frame. In some cases, a video could be recognized with only a single frame like the example of the first row in Fig. \ref{fig:fig1}(a). Current approaches~\cite{ bishay2019tarn,kumar2021few,li2022ta2n} usually extract the spatial features through a TSN ~\cite{wang2016temporal} model. However, they usually consider the high-level spatial features as default but ignore the evenly crucial low-level spatial features, which contain rich texture information. Fusing the low-level spatial features with the high-level ones could compensate and even highlight the low-level semantic features. For the temporal features, we classify them into two categories, long-term and short-term temporal features. Long-term temporal features present the relationship between spatial appearance features of different timestamps, which has also been a hot topic in previous works.
    For instance, the action of ``high jump" in Fig. \ref{fig:fig1}(a) is easily mistaken for ``running" if the feature of jumping into the mat in the last frame is not integrated into all the previous features. Existing methods ~\cite{zhu2018compound,cao2020few,perrett2021temporal} model the long-term temporal features mainly through hand-designed temporal alignment algorithms during the class prototype construction process, which aims to obtain better global features for comparison. 
    On the other hand, short-term temporal features represent the motion characteristics of adjacent frames, i.e., focus on the local temporal relation modeling. For example, in Fig. \ref{fig:fig1}(a), without the short-term temporal information, it is hard to classify whether the action is ``taking the ring out of the box" or ``putting the ring in the box". Nevertheless, we have observed that the short-term temporal modeling remains unexplored for the few-shot action recognition task.

    The critical insight of our work is to provide powerful spatial-temporal features, making it possible to realize effective recognition with a common frame-level class prototype matcher for few-shot action recognition. To this end, we propose a novel method for few-shot action recognition, dubbed \textbf{SloshNet}, a short for \textbf{S}patial, \textbf{lo}ng-term temporal and \textbf{sh}ort-term temporal features integrated \textbf{Net}work. Specifically, to exploit the low-level spatial features, we first design a feature fusion architecture search module (FFAS) to automatically search for the best fusion structure of the low-level and high-level spatial features in different scenarios. Low-level features focus more on texture and structural information, while high-level features focus more on the semantic information, and their combination can enhance the representation of spatial features. Furthermore, based on the extracted spatial appearance features, we introduce a long-term temporal modeling module (LTMM) to model the global temporal relations. Meanwhile, we design another short-term temporal modeling module (STMM) to encode the motion characteristics between the adjacent frame representations and explore the optimal integration of long-term and short-term temporal features. For class prototype matcher, we follow a frame-level method TRX~\cite{perrett2021temporal}, using an attention mechanism to match each query sub-sequence with all sub-sequences in the support set and aggregates this evidence. Fig. \ref{fig:fig1}(b) shows the learned attentions of our SloshNet with TRX, where the attention learned by our method is highly concentrated and more correlated with the action subject, demonstrating the effectiveness of the spatial-temporal modeling of our SloshNet. The main contributions of our work can be summarized as follows: 
	\begin{itemize}
	    \item We propose a simple and effective network named SloshNet for few-shot action recognition, which  integrates spatial, long-term temporal and short-term temporal features.
	    
		\item  We design a feature fusion architecture search module (FFAS) to automatically search for the best combination of the low-level and high-level spatial features.
		
		\item We introduce a long-term temporal modeling module (LTMM) and design a short-term temporal modeling module (STMM) based on the attention mechanism to encode complementary global and local temporal representations.
		
		\item The extensive experiments on four widely-used datasets (Something-Something V2, SSV2~\cite{goyal2017something},  Kinetics~\cite{carreira2017quo}, UCF101~\cite{soomro2012ucf101}, and HMDB51~\cite{kuehne2011hmdb}) demonstrate the effectiveness of our methods. 
		
	\end{itemize}
	
	\section{Related Works}
    
    \subsection{Few-Shot Image Classification}
	The core problem of few-shot image classification is to obtain satisfactory prediction results based on a handful of training samples. Unlike the standard training methods in deep learning, few-shot image classification uses the episodic training paradigm, making a handful of labeled training samples from numerous similar tasks as a surrogate for many labeled training samples. Existing mainstream methods of few-shot classification can mainly be classified as adaptation-based and metric-based. The adaptation-based approaches aim to find a network initialization that can be fine-tuned for unknown tasks using few data, called \textit{gradient by gradient}. The evidence of  adaptation-based approaches can be clearly seen in the cases of MAML~\cite{finn2017model} and Reptile~\cite{nichol2018reptile}. 
	The metric-based approaches aim to find a fixed feature representation in which the target task can be embedded and classified. The effectiveness of this kind of approach has been exemplified in Prototypical Networks~\cite{snell2017prototypical} and Matching Networks~\cite{vinyals2016matching}.  
	In addition, CrossTransformer~\cite{doersch2020crosstransformers} aligns the query and support set based on co-occurrences of image patches that combine metric-based features with task-specific adaptations.
	\subsection{Few-Shot Video Action Recognition}
	Inspired by few-shot image classification, MetaUVFS~\cite{patravali2021unsupervised} apply the adaptation-based method and design an action-appearance aligned meta-adaptation module to model spatial-temporal relations of actions over unsupervised hard-mined episodes. However, the adaptation-based method requires high computational resources and long experimental time, so it is less commonly used in few-shot action recognition compared to the metric-based method.
	In this field, scholars have developed different sub-divisional concerns about the metric-based approach.
	Some of metric-based approaches ~\cite{zhu2018compound,cao2020few,li2022ta2n} focus on hand-designed temporal alignment algorithms during the class prototype construction process. Simultaneously, TRPN~\cite{wang2021semantic} focuses on combining visual and semantic features to increase the uniqueness between similar action classes, and another work~\cite{liu2022task} focuses on frame sampling strategies to avoid omitting critical action information in temporal and spatial dimensions. Moreover, unlike the above methods of prototype matching at the video level, some methods~\cite{perrett2021temporal,thatipelli2022spatio} inspired by CrossTransformer match each query sub-sequence with all sub-sequences in the support set, which can match actions at different speeds and temporal shifts. Our method focuses on modeling spatial-temporal relations based on a handful of labeled data. We can obtain good predictions by feeding rich spatial-temporal features to a common frame-level class prototype matcher like TRX~\cite{perrett2021temporal}.
 	\section{Method}
	   
\begin{figure*} [ht]
		\centering
		\includegraphics[width=\linewidth]{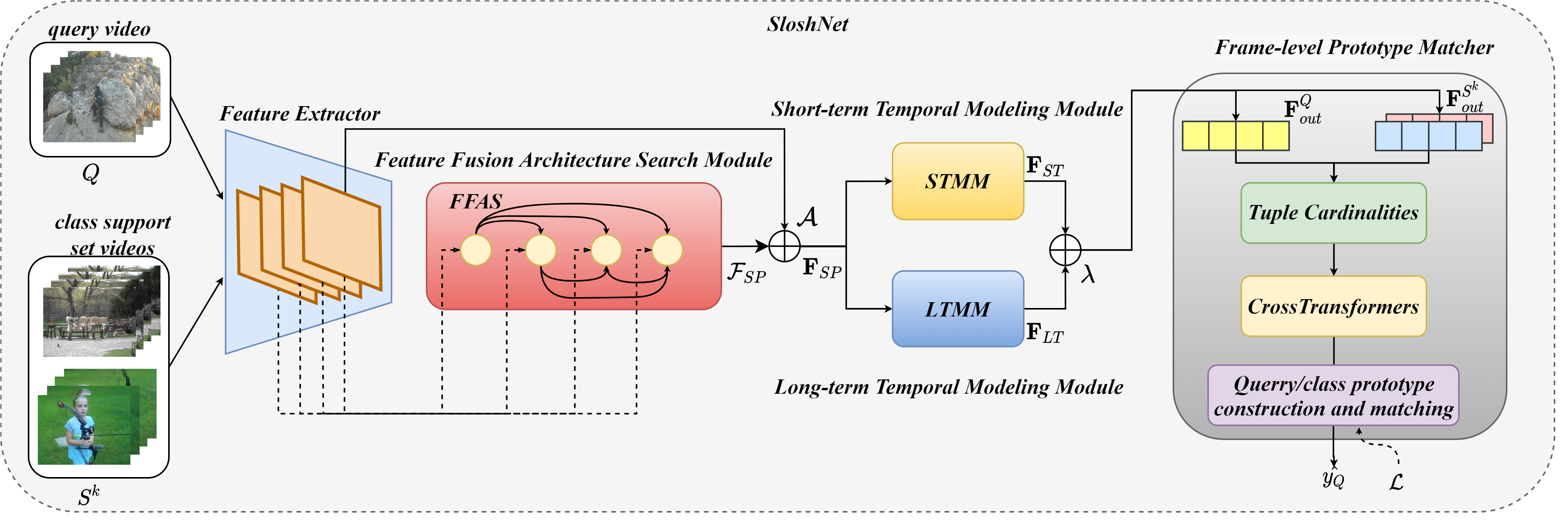}
	\caption{Overview of SloshNet. The spatial fusion feature $\textbf{F}_{SP}$ is obtained by the future fusion architecture search module (FFAS). The long-term temporal feature  $\textbf{F}_{LT}$ is obtained by the long-term temporal modeling module (LTMM). The short-term temporal feature $\textbf{F}_{ST}$ is obtained by the short-term temporal modeling  module  (STMM). The $\widehat{y}_Q$ is the class prediction of the query video and the loss  $\mathcal{L}$ is a standard cross-entropy loss. $\oplus$ indicates element-wise weighted summation with a learnable parameter.}
		\label{fig:pipeline}
	\end{figure*}

	    Fig. \ref{fig:pipeline} illustrates our overall few-shot action recognition framework. The query video $Q$ and the class support set videos $S^k$ passed through the feature extractor, and store the output features of each layer in a feature bank. The features from the feature bank are input into the feature fusion architecture search module (FFAS) to obtain the spatial fusion feature $\textbf{F}_{SP}^{Q}$ , $\textbf{F}_{SP}^{S^k}$.
	    Next, we do the weighted summation of the fused feature and the original last layer feature from the feature bank with a learnable parameter $\gamma$ to obtain the enhanced spatial feature. 
	    Followed previous works ~\cite{yang2020temporal,zhu2021closer,thatipelli2022spatio}, we model the temporal relation after the acquisition of spatial features to obtain better spatial-temporal integration features. 
	    Therefore,  the obtained spatial features will be passed through a long-term temporal modeling module (LTMM) and a short-term temporal modeling module (STMM) to model long-term  and short-term temporal characteristics $\textbf{F}_{LT}^Q, \textbf{F}_{LT}^{S^k},\textbf{F}_{ST}^Q, \textbf{F}_{ST}^{S^k}$ in parallel. Then do the fusion with another learnable parameter, which adaptively fuses the two kinds of temporal features.
	    Finally, the class prediction$\ \widehat{y}_Q$ of the query video $Q$ and loss $\mathcal{L}$ are obtained by a frame-level prototype matcher. Details are shown in the subsequent subsections.
	\subsection{Problem Formulation}
    The few-shot action recognition is considered an N-way, K-shot task. It assigns an unlabeled query video to one of the N classes in the support set, each containing K-labeled videos that were not seen during the training process. We follow an episode training paradigm in line with most previous works~\cite{zhu2018compound,cao2020few,perrett2021temporal}, where episodes are randomly drawn from an extensive data collection, and each episode is seen as a task. In each task, we let $ \ Q=\left\{q_1,q_2,\cdots,q_l\right\}$ denote a query video randomly sampled $l$ frames, and $S_m^k=\left\{s_{m1}^k,s_{m2}^k,\cdots,s_{ml}^k\right\}$ represents the $m^{th}$ video in class $k\ \epsilon\ K$ randomly sampled $l$ frames.

     \subsection{FFAS: Feature Fusion Architecture Search Module}
      The low-level features extracted in the earlier layers of the feature extractor focus more on the structure and texture information, while the high-level features extracted in the last layers focus more on the semantic information. The fusion of them helps improve the spatial representations. Inspired by ~\cite{liu2018darts, ghiasi2019fpn}, we design a feature fusion architecture search module (FFAS). Our goal is to fuse features from different layers output by the feature extractor with an auto-search fusion module, which enables us to find  the best combination of the low-level and high-level spatial characteristics in different scenarios.
    \begin{figure} [ht]
		\centering
		\includegraphics[width=\linewidth,height=0.8\linewidth]{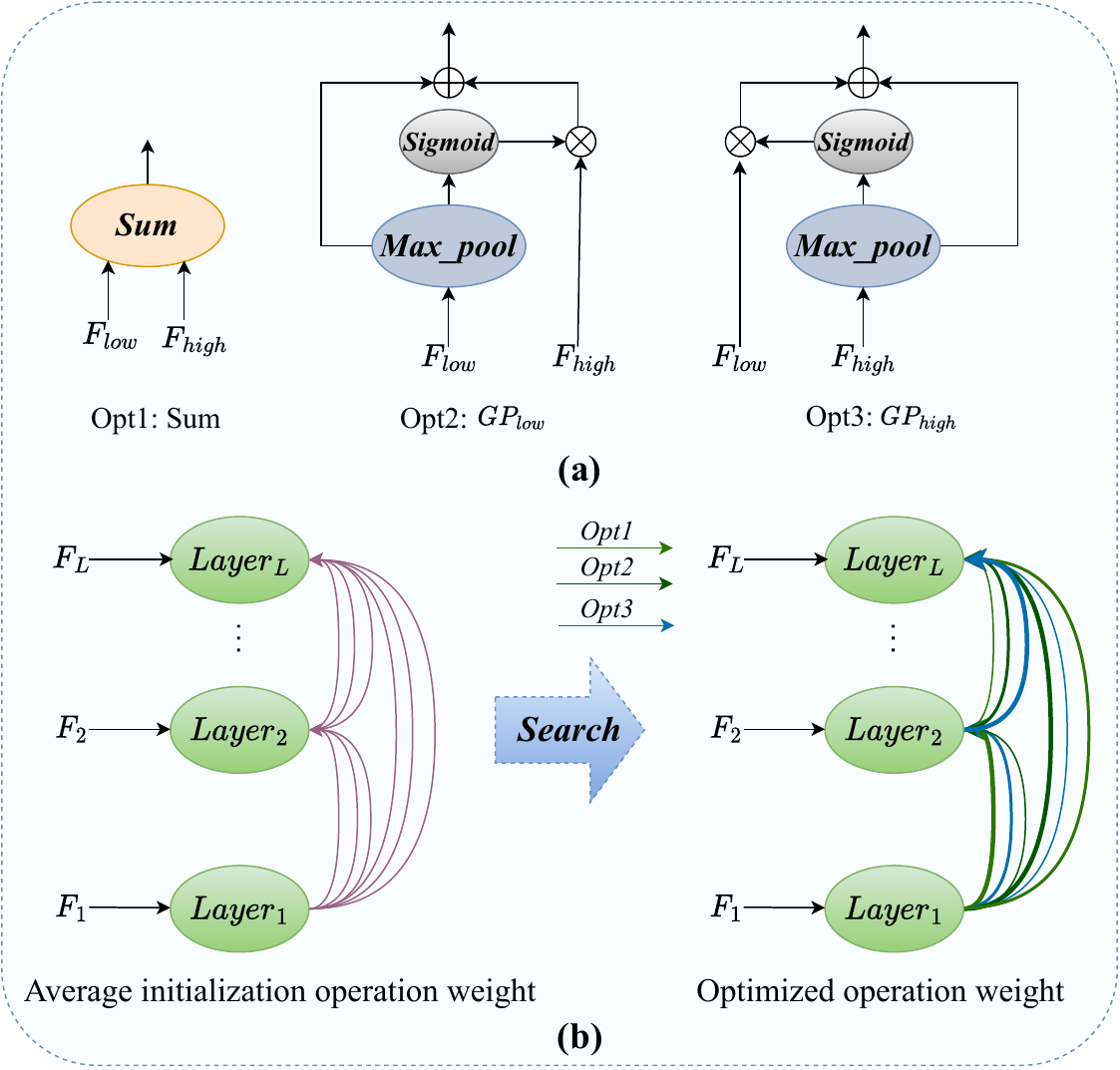}
		\caption{Three feature fusion options, including the sum option and two types of global pooling options, are shown in (a), in which Max\_pool devotes max pooling. (b) shows the feature fusion search process. $\oplus$ indicates element-wise summation and $\otimes$ indicates element-wise product.}
        \vspace{-10pt}
		\label{fig:fig3}
	\end{figure}
    Specifically, we give the features of each layer (total L layers) $\mathcal{F}=\left\{\textbf{F}_1,\cdots,\textbf{F}_i,\cdots,\textbf{F}_L\right\} \  (\textbf{F}_i\in\mathbb{R}^{NT\times {C_i} \times {H_i} \times {W_i}})$ where $N,T,C,H,W$ are the batch size, time, spatial,  height, and width, respectively. To facilitate the subsequent fusion process, we align each layer feature's spatial and channel dimension to the last layer feature, i.e.: $\textbf{F}_i\in\mathbb{R}^{NT\times\ C_i\times\ H_i\times\ W_i}\rightarrow\mathbb{R}^{NT\times\ C\times\ H\times\ W}$  as follows:
    \begin{equation}\label{}
    \textbf{F}_i=Module_{align}\left(\textbf{F}_i\right)
    \end{equation}
    where $Module_{align}$ here is a $3\times3$  convolution layer. After feature alignment, each layer feature will be updated with the fusion of all previous layers' features as:
     \begin{equation}\label{}
     \textbf{F}_j=\sum_{i<j}{{\bar{o}}_{i,j}(\textbf{F}_i,\textbf{F}_j)}
    \end{equation}
    where ${\bar{o}}_{i,j}\left(\textbf{F}_i,\textbf{F}_j\right)$ is the weighted summation result of the features of layer $i$ and $j$ after passing all optional fusion operations. We let the set of these fusion options be indicated as $\mathcal{O}$. In our work, we provide three parameter-free  fusion options $Sum$, $GP_{low}$, and $GP_{high}$, as shown in Fig. \ref{fig:fig3}(a). To make the search space continuous, we assign a weight $\alpha$ to each operation and perform a $softmax$. This search task can be simplified to learning weights $\alpha$, and $o_{i,j}\left(\textbf{F}_i,\textbf{F}_j\right)$ can be calculated as:
     \begin{equation}\label{}
    {\bar{o}}_{i,j}\left(\textbf{F}_i,\textbf{F}_j\right)=\sum_{o\in \mathcal{O}}\frac{exp\left({\alpha_{i,j}}^o\right)}{\sum_{o^\prime\in \mathcal{O}} e x p\left({\alpha_{i,j}}^{o^\prime}\right)}o_{i,j}\left(\textbf{F}_i,\textbf{F}_j\right)
    \end{equation}

    The feature fusion search process is shown in Fig. \ref{fig:fig3}(b), and the weights of each fusion operation are initialized equally. Moreover, the output of the module is the updated features of all layers, denoted as $\mathcal{F}_{SP}=\left\{\textbf{F}_{SP}^1,\cdots,\textbf{F}_{SP}^i,\cdots,\textbf{F}_{SP}^L\right\} \  (\textbf{F}_{SP}^i\in\mathbb{R}^{NT\times C\times H\times W})$.  Finally, we do the weighted summation of the fused feature $\mathcal{F}_{SP}$ and the final feature $\textbf{F}_L$ output by the feature extractor with learnable parameters  $\mathcal{A}=\left\{\gamma_{1},\cdots \gamma_{i},\cdots,\gamma_{L} \right\}( \gamma_{i}\in\left[0,1\right], \sum_{i=1}^L{\gamma _i}<1)$  to obtain the enhanced spatial feature $\textbf{F}_{SP}$, given by
    \begin{equation}\label{}
\mathbf{F}_{SP}=\sum_{i=1}^L{\gamma _i\mathbf{F}_{SP}^{i}}+\left( 1-        \sum_{i=1}^L{\gamma _i} \right) \mathbf{F}_L
    \end{equation}
    \subsection{LTMM: Long-Term Temporal Modeling Module}
    In few-shot action recognition, many objects move over time, so many actions could be classified according to their global temporal contextual information. We employ a long-term temporal modeling module (LTMM) to model the global temporal relations based on the extracted spatial appearance features.
    \begin{figure} [ht]
		\centering
		\includegraphics[width=\linewidth,height=0.6\linewidth]{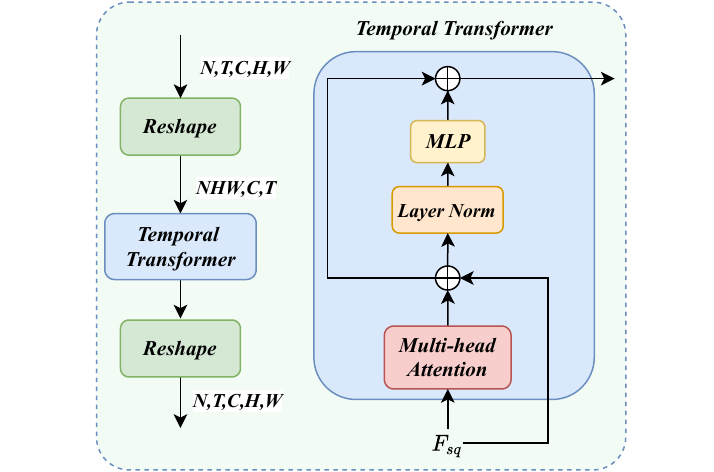}
		\caption{The architecture of the long-term temporal modeling module (LTMM). $\oplus$ denotes element-wise summation.}
		\label{fig:fig4}
          \vspace{-10pt}
	\end{figure}
	   
	 We present the structure of LTMM in Fig. \ref{fig:fig4}. Given a video feature map after spatial enhancement indicated as $\textbf{F}_{SP}{\in\mathbb{R}}^{N\times T\times C\times H\times W}$, it will be reshaped to a sequence shown as $\textbf{F}_{sq}{\in\mathbb{R}}^{NHW\times T\times C}$.  Then,  we let $\textbf{F}_{sq}$ do self-attention in the temporal dimension:
    \begin{equation}\label{}
    \textbf{F}_{sq}=\textbf{F}_{sq}+Module_{att}\left(\textbf{F}_{sq}\right)
    \end{equation}
    where $Module_{att}$ is a $L_t$ layer multi-head attention. The obtained features are then pointed-wise refined by a residual feed-forward network to obtain long-term temporal features $\textbf{F}_{LT}{\in\mathbb{R}}^{NHW\times T\times C}$, given by:
       \begin{equation}\label{}
    \textbf{F}_{LT}=\textbf{F}_{sq}+\varphi(LN(\textbf{F}_{sq}))
    \end{equation}
    where $LN$ denotes the layer normalization and $\varphi$ denotes the multi-layer perceptron. Next, $\textbf{F}_{LT}$ will be reshaped to the original input shape (i.e. [N,T,C,H,W]).
    
    \subsection{STMM: Short-Term Temporal Modeling Module}
    The classification of many action categories requires short-term temporal information, representing the motion characteristics of adjacent frames, and is beneficial for recognizing many temporal-related actions. Therefore, we propose a novel short-term temporal modeling module (STMM) to encode the motion information between adjacent frame representations in the feature level. 
    
    Given a video feature map after spatial enhancement $\textbf{F}_{SP}{\in\mathbb{R}}^{N\times T\times H\times W\times C}$, we obtain query-key-value triplets using learnable weights $W_1, W_2, W_3\in\mathbb{R}^{D\times D}$, 
     \begin{equation}\label{}
     \textbf{F}^q=\textbf{F}_{SP}\textbf{W}_1,\ \ \textbf{F}^k=\textbf{F}_{SP}\textbf{W}_2,\ \ \textbf{F}^v=\textbf{F}_{SP}\textbf{W}_3\ 
     \end{equation}
    Next, we reshape $\textbf{F}^q,\ \textbf{F}^k{\in\mathbb{R}}^{NTr\times C/r\times H\times W}$ to reduce the channels by a factor of $r$ to ease the computing cost and leverage two channel-wise $3\times3$ convolution layers $\textbf{K}^q$, $\textbf{K}^k$ on $\textbf{F}^q$ and $\textbf{F}^k$, given by
    \begin{equation}\label{}
    \textbf{F}^q=\sum_{i,j} \textbf{K}_{c,i,j}^q\textbf{F}_{c,h+i,w+j}^q ,\ \  \textbf{F}^k=\sum_{i,j} \textbf{K}_{c,i,j}^k\textbf{F}_{c,h+i,w+j}^k
    \end{equation}
    where $c,h,w$ represent the channel and two spatial dimensions of the feature map. $\textbf{K}_{c,i,j}^q$ and $\textbf{K}_{c,i,j}^k$ indicate the $c^{th}$ filter, with the subscripts$\ i, j\epsilon\left\{-1,0,1\right\}$ to show the spatial indices of the kernel. Next, we do the staggered subtraction over the temporal dimension between $\textbf{F}^q$ and $\textbf{F}^k$ to obtain the motion information in feature level, i.e. $\textbf{F}_t^q$ and $\textbf{F}_{t+1}^k$. Formally,
    \begin{equation}\label{}
    \textbf{M}=Concat\left(\left(\textbf{F}_1^q-\textbf{F}_2^k\right),\cdots,\left(\textbf{F}_t^q-\textbf{F}_{t+1}^k\right)\right)
    \end{equation}
    where for the $L$ frames video $\left(1\le t\le L-1\right)$. The temporal dimension of the motion representation $\textbf{M}$ is $T-1$, so we use zero to represent the motion information of the last time step to help $\textbf{M}$ keep the temporal size compatible with the input feature maps. Then, the $\textbf{M}$ is reshaped to the shape of original input features $\textbf{M}{\in\mathbb{R}}^{N\times T\times H\times W\times C}$ to restore the number of channels to $C$. In the end, a feed-forward network (FFN) is applied on the motion attention \textbf{M} and the final output of STMM is obtained as:
    \begin{equation}\label{}
    \textbf{F}_{ST}=Sigmoid\left(MLP\left(GELU\left(MLP\left(\textbf{M}\right)\right)\right)\right)\textbf{F}^v
    \end{equation}
    
\begin{figure} [ht]
		\centering
		\includegraphics[width=\linewidth,height=0.87\linewidth]{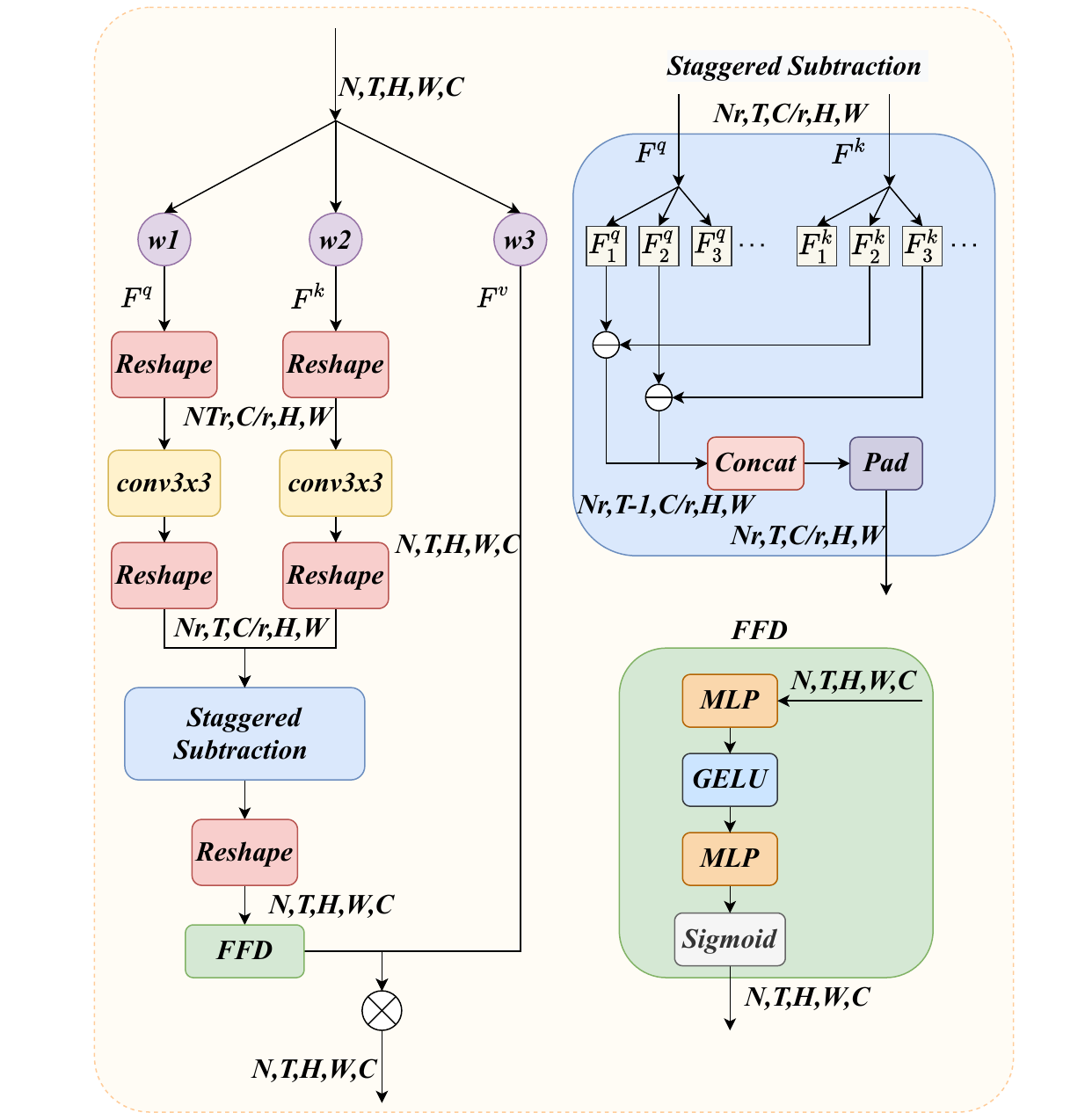}
		\caption{The architecture of the short-term temporal modeling module (STMM). $\ominus$ indicates element-wise subtraction, and $\otimes$ shows the element-wise product.}
		\label{fig:fig5}
          \vspace{-10pt}
	\end{figure}

    The structure of STMM is shown in Fig. \ref{fig:fig5}. Finally, we do the weighted summation of the short-term temporal features $\textbf{F}_{ST}$ and the long-term temporal features $\textbf{F}_{LT}$ with a learnable parameter $\lambda\epsilon\left[0,1\right]$, given by
    \begin{equation}\label{}
    \textbf{F}_{out} = \left(1-\lambda\right) \textbf{F}_{ST}+\lambda \textbf{F}_{LT}
    \end{equation}
	
        \subsection{Class Prototype Matcher}
    
    Frame-level prototype construction and matching facilitate fine-grained classification of few-shot action recognition. In our work, we want to show that good predictions can be obtained by feeding rich spatial-temporal features to a common frame-level class prototype matcher. We followed the TRX~\cite{perrett2021temporal}, a common frame-level prototype matcher, which matched each query sub-sequence with all sub-sequences in the support set to construct a query-specific class prototype.
    
    Specifically, we first construct a frame-level feature representation of the video. $r_i{\in\mathbb{R}}^D$ denotes the $i^{th}$ frame representation and a sequence representation between the $i^{th}$ frame and the $j^{th}$ frame $(\omega=2)$ is shown as $\left(r_i,r_j\right){\in\mathbb{R}}^{2D}$, where $1\le i\le j\le l$, and so on. For any tuple $t\in\mathrm{\Pi}^\omega (\omega\epsilon\Omega)$, aggregate all possible sub-sequences in the support video $S_{mt}^k{\in\mathbb{R}}^{\omega D}$ of an action class to compute a query-specific class prototype, where the aggregation weights are based on the cross-attention of the query sub-sequence and the support class sub-sequence. Let the query embedding indicate as $\textbf{q}_t{\in\mathbb{R}}^{D^\prime}$ and the query-specific class prototype denote as ${\textbf{u}^k}_t{\in\mathbb{R}}^{D^\prime}$. Then, the distance between a query video $Q$ and a class in support set $\textbf{S}^k$ over multiple cardinalities $\Omega$ can be calculated as:
     \begin{equation}\label{}
    \textbf{D}\left(Q,\textbf{S}^k\right) = \sum_{\omega\epsilon\Omega}\frac{1}{\left|\mathrm{\Pi}^\omega\right|} \sum_{t\in\mathrm{\Pi}^\omega}{||\textbf{q}_t-{\textbf{u}^k}_t||}
    \end{equation}
    The distance $\textbf{D}(\cdot,\cdot)$ is minimized by a standard cross-entropy loss from the query video to its ground-truth class. During the inference, the query is assigned the class closest to the query with $\textbf{D}$, i.e., $argmin(\textbf{D})$.
    
\section{Experiments}
\subsection{Experimental Setup} 

\subsubsection{Network Architectures} We use the ResNet-50 as the feature extractor with ImageNet pre-trained weights~\cite{deng2009imagenet}. In FFAS, we automatically search for the best combination of the four layers in ResNet-50 and the weights of the three optional operations are initialized equally in each layer. We use the $3\times3$ convolution layer as the $Module_{align}$, the spatial self-attention as the $f_{fuse}$ in FFAS and two layers multi-head attention as the $Module_{att}$ in 
LTMM. $r$ in STMM is set to 16. The initial weight of the learnable parameter $\mathcal{A}$ and $\lambda$ is set to [0.1, 0.1, 0.1, 0.1] and 0.5, respectively.  In frame-level class prototype matcher, we set $D^\prime$ = 1152, $\Omega=\left\{1\right\}$ for spatial-related datasets, and $\Omega=\left\{1,2\right\}$ for temporal-related dataset. 

\subsubsection{Training and Inference} We uniformly sampled 8 frames ($l$=8) of a video as the input augmented with random horizontal flipping and $224\times224$ crops in training, while only a center crop in inference. For training, SSV2 were randomly sampled 100,000 training episodes with an initial learning rate of ${10}^{-4}$, and the other datasets were randomly sampled 10,000 training episodes with an initial learning rate of ${10}^{-3}$. Moreover, we used the SGD optimizer with the multi-step scheduler for our framework. For inference, we reported the average results over 10,000 tasks randomly selected from the test sets in all datasets.

\subsection{Results}
\subsubsection{Results on Spatial-Related Datasets} For the spatial-related datasets, the recognition of actions depends mainly on the background information and a small part on the temporal information. So in the experiments of these datasets, we set $\Omega=\left\{1\right\}$, making it more focused on background information during the class prototype construction and matching. Also, since we have modeled the long-term and short-term temporal relations at the feature level, each frame feature has an intrinsic temporal representation. The state-of-the-art comparison for the 5-way 5-shot action recognition task of three spatial-related datasets, including Kinetics, HMDB51, and UCF101, was shown in Tab. \ref{5w-5s spatial}. On all three datasets, we achieve the new state-of-the-art results. Taking the Kinetics as an instance, compared to our baseline TRX~\cite{perrett2021temporal}, we bring a 1.1$\%$ performance improvement demonstrating the effectiveness of our spatial-temporal relation modeling. Meanwhile, compared to the similar method STRM~\cite{thatipelli2022spatio} that focused on spatial-temporal modeling but lacked short-term temporal modeling, our approach brings a 0.5$\%$ accuracy improvement showing the significance of characteristics between adjacent frame representations.
Similarly, our SloshNet achieves the best performance in HMDB51 and UCF101.
    
\begin{table}[t]
		\setlength\tabcolsep{2pt}
		\centering
		\scalebox{0.90}{\begin{tabular}{cccc}
				\hline
				Methods                     & Kinetics        & HMDB & UCF \\ \hline\hline
				CMN \cite{zhu2018compound}               & 78.9      & -  &  - \\ \hline
                ProtoNet & \multirow{2}{*}{64.5} & \multirow{2}{*}{54.2} & \multirow{2}{*}{78.7} \\
				\cite{snell2017prototypical}                 &                     &                       &   \\ \hline
				TARN & \multirow{2}{*}{78.5} & \multirow{2}{*}{-} & \multirow{2}{*}{-} \\
				\cite{bishay2019tarn}                 &                     &                       &   \\ \hline
				ARN \cite{zhang2020few}   & 82.4  & 60.6  & 83.1 \\ \hline
			
				OTAM \cite{cao2020few}            & 85.8  & -  & - \\\hline
				HF-AR \cite{kumar2021few}    & -  & 62.2  & 86.4 \\ \hline
				TRX  ~\cite{perrett2021temporal}       & 85.9  & 75.6  & 96.1 \\ \hline
				TA2N \cite{li2022ta2n}             & 85.9  & 75.6  & 96.1 \\ \hline
				STRM\newline{}~\cite{thatipelli2022spatio}  & 86.5  & 77.3  & 96.9 \\ \hline
				\textbf{SloshNet}   &\textbf{87.0}    & \textbf{77.5}  & \textbf{97.1} \\ \hline
		\end{tabular}}
		\caption{State-of-the-art comparison on the 5-way 5-shot spatial-related  benchmarks of Kinetics, HMDB51, and UCF101. } 
  		\label{5w-5s spatial}
            \vspace{-10pt}
	\end{table}

\begin{table}[t]
		\setlength\tabcolsep{2pt}
		\centering
		\scalebox{0.90}{\begin{tabular}{ccc}
				\hline
	    
	    & \multicolumn{2}{c}{SSV2} \\ 
        \multirow{-2}{*}{Methods}  & 1-shot & 5-shot \\  \hline \hline
		 ProtoNet & \multirow{2}{*}{39.3} & \multirow{2}{*}{52.0} \\
				\cite{snell2017prototypical}                 &                     & \\ \hline
		 OTAM \cite{cao2020few}            & 42.8  & 52.3 \\\hline
		 HF-AR \cite{kumar2021few}         & 43.1  & 55.1 \\ \hline
		 PAL \cite{zhu2021few}                & 46.4  & 62.6 \\ \hline
		 TRX  ~\cite{perrett2021temporal}       & 42.0  & 64.6 \\ \hline
    	 STRM  \cite{thatipelli2022spatio}  & 43.5  & 66.0* (68.1) \\ \hline
    	 \textbf{SloshNet}  & \textbf{46.5}    & \textbf{68.3} \\ \hline
		\end{tabular}}
		\caption{State-of-the-art comparison on the 5-way  temporal-motion-related  benchmark of SSV2. * refers to our re-implementation and () refers to the reported result.} 
  		\label{5w-5s temporal}
	\end{table}
	
\subsubsection{Results on Temporal-Related Dataset} For the temporal-related dataset SSV2, the key to action recognition is long-term and short-term temporal information. Therefore, we set $\Omega=\left\{1,2\right\}$ to reinforce long-term and short-term temporal relation modeling both during feature-level and class prototype construction and matching process. The state-of-the-art comparison for the 5-way 1-shot and 5-way 5-shot action recognition tasks of the temporal-related dataset SSV2 was shown in Tab. \ref{5w-5s temporal}. Compared to the best existing method STRM~\cite{thatipelli2022spatio} in SSV2, SloshNet has a large improvement of 3.0$\%$ on 5-way 1-shot task and 2.3$\%$ on 5-way 5-shot task.

\subsection{Ablation Study} 
\subsubsection{Impact of Proposed Modules} 
\begin{figure*} [ht] 
		\centering
		\includegraphics[width=\linewidth ,height=0.25\linewidth]{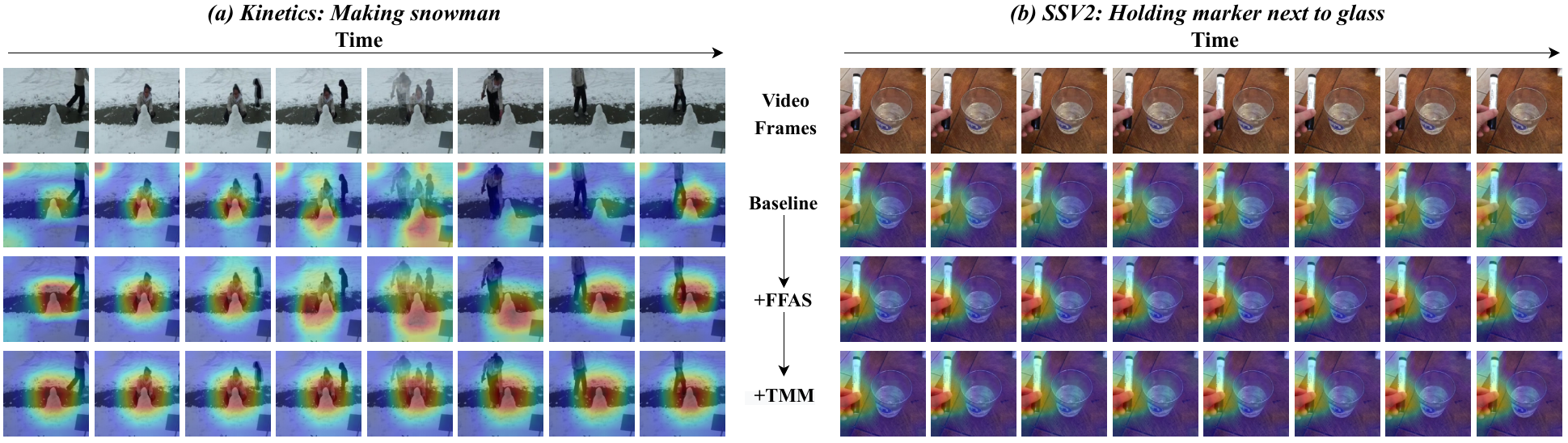}
	\caption{Attention visualization of our SloshNet on two examples. From top to bottom, we gradually integrate the impact of our contribution, in which FFAS indicates feature fusion architecture search and TMM denotes the combination of short-term and long-term temporal relation modeling.}
		\label{fig:att}
          \vspace{-10pt}
\end{figure*}

To validate the contributions of each module in the SloshNet, we experiment on the 5-way 1-shot task of SSV2 ($\Omega=\left\{1,2\right\}$) to ablate our proposed components in Tab. ~\ref{proposed modules}. The spatial representation enhancement module FFAS brings about a 0.8$\%$ accuracy improvement. LTMM and STMM combined as the temporal modeling module (TMM) bring a 1.6$\%$ gain. When combining FFAS and TMM, we can learn spatial and temporal features together and achieve the best accuracy, with the gain of 4.5$\%$ over the baseline. Meanwhile, we also provide the attention visualization of our SloshNet in Fig. ~\ref{fig:att}, which gradually integrates the impact of our contribution. After integrating FFAS (third row), our framework enhances the feature spatial representation, which helps concentrate on relevant objects in a single video frame, e.g., the frames in (a) and (b) reduce attention on the background and extraneous objects.  Furthermore, after integrating TMM (fourth row), including LTMM and STMM, our framework enhances the feature temporal relation, which lets our SloshNet highly correlate with the action subject, e.g., the fourth and eighth frame from the left in (a) has a better concentration on the snowman, and the frames in (b) has more detailed attention extended to the marker pen. 
\begin{table}[tb]
		\setlength\tabcolsep{10pt}
		\centering
				\scalebox{0.90}{\begin{tabular}{cccc}
				\hline
	    	FFAS            & LTMM  & STMM & Acc \\ \hline\hline
           $\times$ 	 &  $\times$     & $\times$   & 42.0 \\ \hline  
	        $\checkmark$	  &  $\times$     & $\times$  & 42.8 \\ \hline
    	$\times$  &  $\checkmark$    &   $\checkmark$& 43.6 \\ \hline

    $\checkmark$	 &$\checkmark$     & $\checkmark$ & \textbf{46.5} \\ \hline
		\end{tabular}}
		\caption{The impact of proposed modules.} 
  		\label{proposed modules}
            \vspace{-14.5pt}
	\end{table}

\subsubsection{Impact of Options and Architecture Search Mechanism in FFAS} 
There are three parameter-free feature fusion options in FFAS: $Sum$, $GP_{low}$, and $GP_{high}$. We conduct 5-way 1-shot task experiments on SSV2 ($\Omega=\left\{1,2\right\}$) to explore the impact of the individual option, combination of options, and architecture search mechanism in FFAS shown in Tab. \ref{Options in FFAS}. Perform $Sum$, $GP_{low}$, $GP_{high}$ options individually, which outperforms baseline by 0.6$\%$, 0.3$\%$, 0.9$\%$, respectively. Above three options are performed simultaneously and summed with equal weights as output, bringing a 2.6$\%$ accuracy improvement. When using the architecture search module under the three feature fusion options, we can yield the best result, bringing a 2.9$\%$ accuracy improvement.
\begin{table}[th]
		\setlength\tabcolsep{3pt}
		\centering
				\scalebox{0.90}{\begin{tabular}{ccccc}
				\hline
	    	$Sum$  &  $GP_{low}$ & $GP_{high}$ & Architecture Search & Acc \\ \hline\hline
	    	$\times$  	 &  $\times$     & $\times$ & $\times$ & 43.6 \\ \hline
          $\checkmark$ 	 &  $\times$     & $\times$ & $\times$ & 44.2 \\ \hline 
	         $\times$	  & $\checkmark$    & $\times$& $\times$  &43.9\\ \hline
    	$\times$  &    $\times$    &   $\checkmark$ &  $\times$& 44.5 \\ \hline
	$\checkmark$  &    $\checkmark$    &   $\checkmark$ &  $\times$& 45.2 \\ \hline
    $\checkmark$  &    $\checkmark$    &   $\checkmark$ &  $\checkmark$& \textbf{46.5}\\ \hline
		\end{tabular}}
		\caption{The impact of options and architecture search mechanism in FFAS.} 
  		\label{Options in FFAS}
            \vspace{-10pt}
	\end{table}
\subsubsection{Impact of Temporal Modeling Integration} 
We also discuss the impact of temporal modeling integration shown in Tab. \ref{temporal modeling}. We conclude that neither short-term nor long-term temporal relation modeling can fully obtain the representation of temporal features. Moreover, compared to the concatenation and parallel connection summation, parallel connection weighted summation with a learnable parameter $\lambda$ is the most effective way to integrate long-term and short-term temporal features, which brings a 3.7$\%$ accuracy improvement to the no any temporal modeling one.

\begin{table}[th]
		\setlength\tabcolsep{10pt}
		\centering
				\scalebox{0.90}{\begin{tabular}{cc}
				\hline
	        Temporal Modeling Integration	 & Acc \\ \hline\hline
          no any temporal modeling & 42.8 \\ \hline  
	        only STMM	  & 43.3 \\ \hline
    	only LTMM  &   43.7 \\ \hline
        	LTMM + STMM  &   43.9 \\ \hline
        	STMM + LTMM  &   44.8 \\ \hline
        	STMM // LTMM  &   45.7 \\ \hline
            STMM $\oplus$  LTMM  &   \textbf{46.5} \\ \hline
		\end{tabular}}
		\caption{The impact of temporal modeling integration. ``+'' indicates concatenation, ``//'' indicates parallel connection summation, and $\oplus$ indicates parallel connection weighted summation with a learnable parameter $\lambda$.}
  		\label{temporal modeling}
	\end{table}
\subsubsection{Impact of Varying Cadinalities} 
Tab. \ref{cadinalities} shows the impact of varying cardinalities given temporal relation modeling during the class prototype construction and matching process. As part of our experiments, we perform 5-way 5-shot task on Kinetics and 5-way 1-shot task on SSV2. 
On both datasets, we then evaluate each cardinality of $\Omega\ \epsilon\left\{1,2,3\right\}$ independently and all their combinations. As for the spatial-related dateset Kinetics, we find $\Omega=\left\{1\right\}$ can achieve the best accuracy of 87.0$\%$, which makes it more focused on background information during the class prototype construction and matching. Furthermore, for the temporal-related dataset SSV2, we set $\Omega=\left\{1,2\right\}$ to reinforce temporal relation modeling during this process, which can get the best accuracy of 46.5$\%$. In fact, for all datasets, high accuracy can be acquired at $\Omega=\left\{1\right\}$, which proves that we already have rich spatial-temporal representation at the feature level. Taken together, these results suggest that if strong enough spatial-temporal feature representations are extracted, the matching process could be simplified a lot.



\begin{table}[tb]
		\setlength\tabcolsep{2pt}
		\centering
				\scalebox{0.90}{\begin{tabular}{cccccccc}
				\hline
	    Cadinalities ($\Omega$)    &$\left\{1\right\}$  & $\left\{2\right\}$ & $\left\{3\right\}$ &$\left\{1,2\right\}$ &$\left\{1,3\right\}$ &$\left\{2,3\right\}$ &$\left\{1,2,3\right\}$ \\ \hline\hline
           Kinetics 	 & \textbf{87.0} & 86.7 & 86.5  & 86.7 & 86.6 & 86.5 & 86.8 \\ \hline  
	        SSV2	  &  46.3 & 46.1 & 45.9 &\textbf{46.5} &46.3 &46.0 &46.3 \\ \hline
   
		\end{tabular}}
		\caption{The impact of varying the cardinallities on Kinetics and SSV2. Comparing all values of $\Omega$ for our SloshNet.  } 
  		\label{cadinalities}
                \vspace{-10pt}
	\end{table}
	
\section{Conclusion}
This paper presents a few-shot action recognition framework, SloshNet, which integrates spatial, long-term temporal, and short-term temporal features into a unified framework. A feature fusion architecture search module (FFAS) is proposed to automatically search for the best combination of the low-level and high-level spatial features to enhance feature spatial representation. A long-term temporal modeling module (LTMM) is introduced to model the global temporal relations based on the extracted spatial appearance features, and a short-term temporal modeling module (STMM) is proposed to encode the motion characteristics between adjacent frame representations. Comprehensive experiments demonstrate the effectiveness of every module and the whole framework.    
\section{Acknowledgements}	
This work is partly supported by the following grant: Key R\&D Program of Zhejiang (No.2022C03126).

\bibliography{aaai23}
\clearpage

\section{Supplementary materials}
 \subsection{A: Datasets}
    \subsubsection{Datasets} We evaluate the performance of our method on four datasets, which can be classified into two categories: 1) spatial-related datasets, including Kinetics~\cite{carreira2017quo}, HMDB51~\cite{kuehne2011hmdb}, and UCF101~\cite{soomro2012ucf101}. 2) temporal-related dataset, Something-SomethingV2~\cite{goyal2017something}. For SSV2 and Kinetics, we used the splits provided by \cite{cao2020few} and \cite{zhu2020label}, where 100 classes were selected and divided into 64/12/24 action classes in training/validation/testing. Meanwhile, for UCF101 and HMDB51, we evaluate our method on the splits provided by ~\cite{zhang2020few}.

	\subsection{B: Results on 5-way 1-shot Spatial-related Benchmarks}
\begin{figure} [h]
		\centering
		\includegraphics[width=\linewidth]{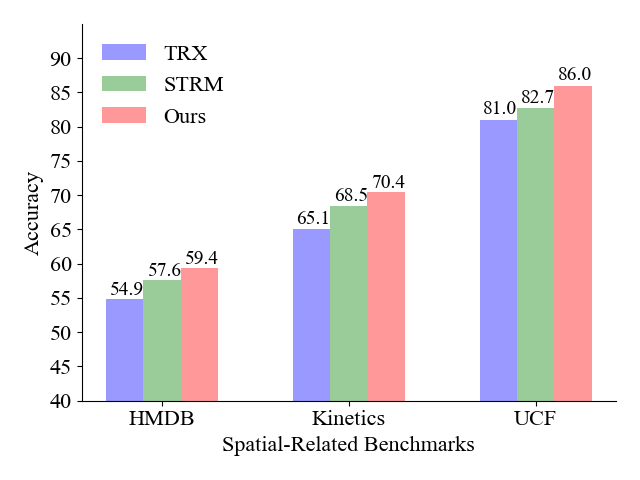}
		\caption{Results on 5-way 1-shot spatial-related benchmarks using the similar frame-level class prototype matcher.}
		\label{fig:fig6}
	\end{figure}

	The results on 5-way 1-shot spatial-related benchmarks was shown in Fig.  \ref{fig:fig6}. As shown in this figure, our SloshNet achieves the most advanced results on the 5-way 1-shot spatial-related benchmarks using the similar frame-level class prototype matcher. Specifically, compared to our baseline TRX~\cite{perrett2021temporal}, we bring 4.5$\%$, 5.3$\%$, and 5.0$\%$ performance improvements on three benchmarks, respectively, which demonstrates the effectiveness of our spatial-temporal relation modeling. Meanwhile, compared to the similar method STRM~\cite{thatipelli2022spatio} that focused on spatial-temporal modeling but lacked short-term temporal modeling, our approach brings  1.8$\%$, 1.9$\%$, and 2.3$\%$ accuracy improvements on three benchmarks, respectively, which emphasizes the significance of characteristics between adjacent frame representations.
	
	\subsection{C: Attention Visualization of SloshNet on Four Benchmarks}
\begin{figure*} [htb]
		\centering
		\includegraphics[width=\linewidth,height=0.5\linewidth]{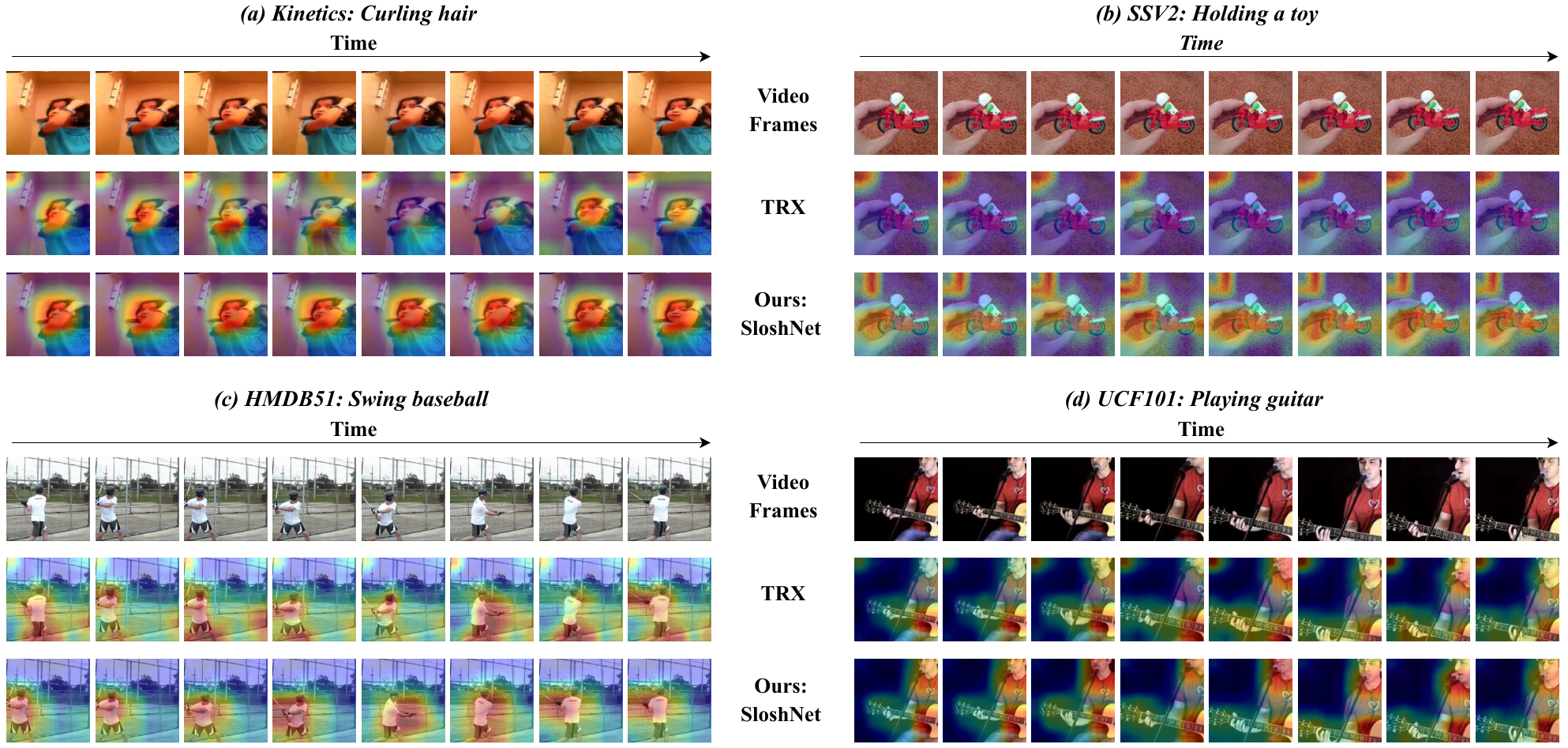}
	\caption{Attention visualization of the recent work TRX~\cite{perrett2021temporal} and our SloshNet on four benchmarks.}
		\label{fig:big_att}
	\end{figure*}

	We also provide the learned attention visualization of our SloshNet with TRX~\cite{perrett2021temporal} on four benchmarks in Fig. \ref{fig:big_att}. TRX (second row) has many unnecessary distractions or even unrelated attentions, e.g., the frames in (b) focus almost entirely on the background, and the others focus much on the extraneous objects. In contrast, our method SloshNet (third row) highly correlates with the action subject, e.g., the frames in (a) mainly focus on the head of the little girl; the frames in (b) mainly concentrate on the hand holding the toy; the frames in (c) attach much attention on the person wielding the baseball bat; the frames in (d) emphasize the person holding the guitar. 
	\subsection{D: Search Results Visualization of FFAS on Four Benchmarks}
\begin{figure*} [htb]
		\centering
		\includegraphics[width=\linewidth,height=0.65\linewidth]{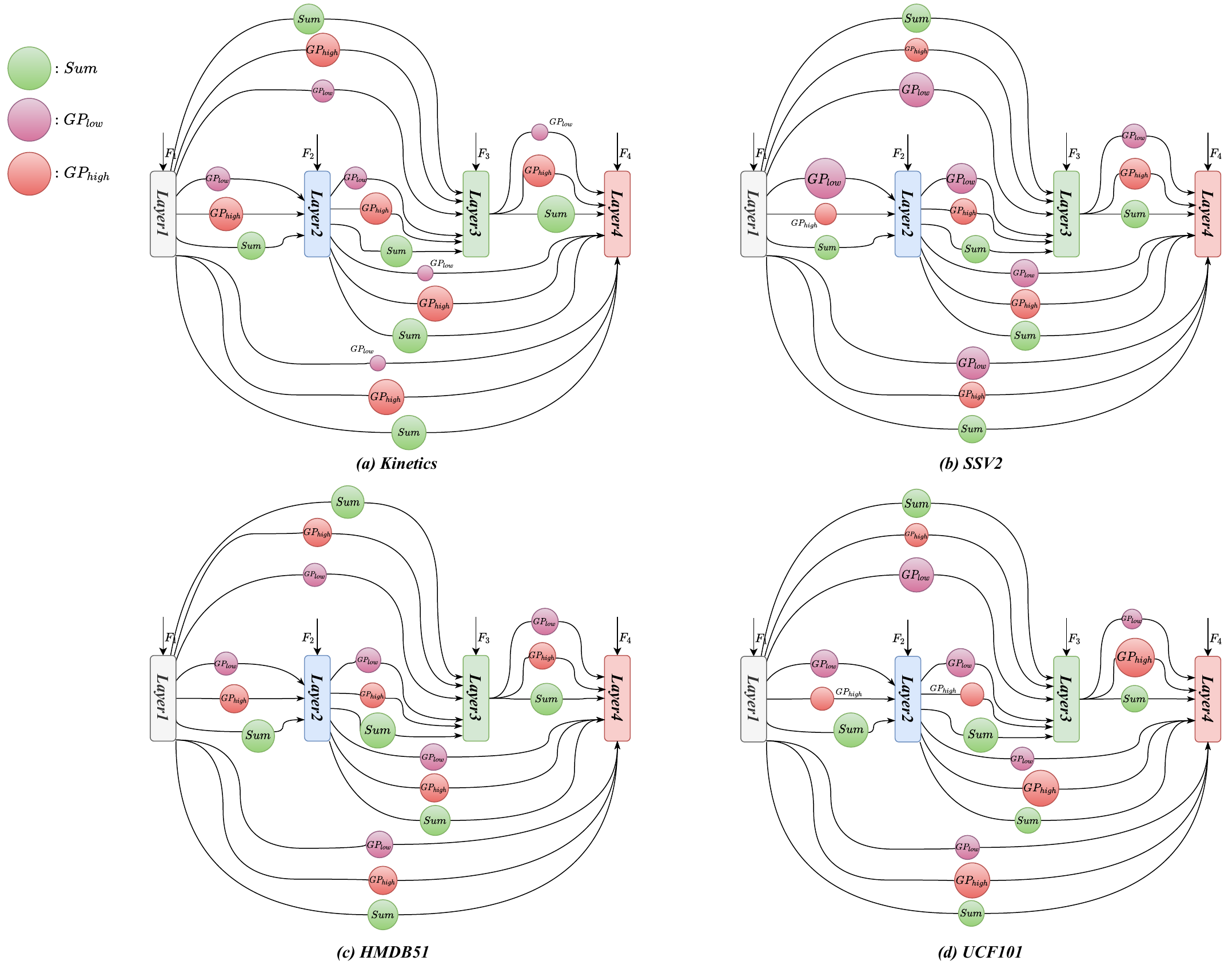}
	\caption{Search results visualization of FFAS on four benchmarks}
		\label{fig:FFAS}
	\end{figure*}

    We have visualized the search results of FFAS on the four benchmarks, as shown in the Fig. \ref{fig:FFAS}. According to the diagram, we visualize each operation as a circle. Specifically, the ratio of the search results, i.e., the weights $\alpha$, between operations and the ratio of the radii between the circles, are in proportion. FFAS can search automatically in different scenarios, making the results optimal for that scenario, just as the search results perform differently in distinct datasets in the graph.
   
\end{document}